\documentclass{article} 
\usepackage{iclr2027_conference,times}

\usepackage{amsmath,amsfonts,bm}

\def\figref#1{Fig.~\ref{#1}}
\def\Figref#1{Figure~\ref{#1}}

\def\secref#1{Sec.~\ref{#1}}

\def\tabref#1{Tab.~\ref{#1}}
\def\Tabref#1{Table~\ref{#1}}
\def\appref#1{App.~\ref{#1}}

\def\eqref#1{(\ref{#1})}

\def\1{\bm{1}}

\DeclareMathAlphabet{\mathsfit}{\encodingdefault}{\sfdefault}{m}{sl}
\SetMathAlphabet{\mathsfit}{bold}{\encodingdefault}{\sfdefault}{bx}{n}

\usepackage{hyperref}
\usepackage{url}

\usepackage{soul}
\usepackage{graphicx}
\usepackage{multirow}
\usepackage{makecell} 

\usepackage{scalerel}
\usepackage{graphicx}
\usepackage{wrapfig}
\usepackage{subcaption}
\usepackage{booktabs}
\usepackage{tabularx}

\title{Do as the Romans Do: Learning Universal \\ Behaviors from Heterogeneous Agents}

\author{Caleb Chang$^1$, Davin Win Kyi$^1$, Natasha Jaques$^{*1}$, Karen Leung$^{*1,2}$ \\
  $^1$ University of Washington, $^2$ NVIDIA \\
  $^*$ indicates equal advising. 
  Correspondence: \texttt{cchang26@uw.edu}}

\iclrfinalcopy 
\begin{document}

\maketitle

\begin{abstract}
Humans often acquire new skills by observing others, since observed behaviors implicitly reveal how to act reasonably in an environment. However, observations drawn from a heterogeneous population introduce conflicting behavioral signals, making it difficult to determine which behaviors are worth imitating. We address this challenge with General Reward Inference and Disentanglement (GRID), a social learning method that extracts universally useful behaviors from a heterogeneous population of demonstrators pursuing different goals. GRID decomposes per-agent reward functions into a general reward, capturing behaviors shared across all agents, and specific rewards, capturing individual preferences and objectives, through an information bottleneck. Training exclusively on the general reward provides a new paradigm of generalist pretraining. It yields a generalist agent that internalizes universal environmental competencies, such as safety and basic task proficiency, without the mode-averaging bias that afflicts standard learning from demonstration techniques. This generalist serves as a strong prior for fine-tuning to downstream tasks, including preferences unseen during training. Experiments across a synthetic basis function decomposition, multi-agent Craftax, continuous control tasks (MuJoCo Gym) and an autonomous driving simulator (Highway-Env) confirm that GRID successfully disentangles reward structure in a semantically meaningful way, outperforms standard learning from demonstration baselines, and enables more efficient and stable specialization.

\end{abstract}

\section{Introduction}

``When in Rome, do as the Romans do\footnote{The proverb advises us to follow the customs and traditions of the place we are visiting.}''---but \textit{which} Roman should we follow? Observed behaviors of individuals implicitly reveal information about how one should act in the environment, yet not all agents follow the same objective, and blindly imitating any one of them can be dangerous or counterproductive. Consider a robot placed among a crowd at a traffic intersection: different pedestrians have different destinations, so they cross the street in different directions---but everyone respects the crosswalk signal. The robot must infer that the crosswalk status is the relevant shared cue, while individual destinations and the behaviors that pursue them are not. Or consider a robot navigating a building where a patch of floor along the shortest path is slippery. Humans will naturally avoid this patch, taking a longer route regardless of their destination. An intelligent agent does not need to know any individual's goal to extract something valuable: the avoidance behavior is universal across all agents. By identifying what every agent consistently does, independent of their individual objectives, a learning agent can infer what is broadly important and adopt those behaviors itself.

The challenge is extracting these shared signals from a heterogeneous population of demonstrators. Standard learning from demonstration (LfD) is vulnerable to potentially conflicting, multi-modal demonstrations and demands additional mechanisms to address variations \citep{SerajLeeEtAl2024}.  Inverse reinforcement learning (IRL), a subset of LfD, recovers an unknown reward function. However, when a population pursues heterogeneous objectives, standard IRL algorithms tend to average conflicting goals together, producing a suboptimal policy that fails to achieve any specific task effectively \citep{PoddarWanEtAl2024}. In contrast, reinforcement learning (RL) learns directly from environmental feedback through a known reward function \citep{SuttonBarto2018}. However, RL is typically sample-inefficient and brittle to environmental changes, leading to poor generalization. In short, neither RL nor standard LfD gracefully handles learning from a population of agents pursuing different goals, even though these agents may reveal extremely valuable information.

This raises a fundamental question: \emph{What should an agent learn from such a population}? We posit that the most valuable signals are those \emph{shared across agents}---behaviors that are universal precisely because they reflect general competence, safety, and environmental norms, rather than any individual's specific objective. All pedestrians obey the crosswalk signal regardless of destination. All skilled drivers maintain safe following distances regardless of where they are headed. All navigators avoid the slippery floor regardless of their target. These shared behaviors are the ``Roman customs'' worth adopting---general norms of the environment that any capable agent implicitly follows.

Social learning, a combination of RL and IRL, moves in this direction. An agent observes other agents in the environment to infer relevant environment information and objectives, and uses that knowledge to inform its own behavior through its own interactive experiences with the environment. Prior works \citep{FilosLyleEtAl2021, AbdulhaiJaquesEtAl2023} explicitly disentangle shared features of the environment from agent-specific goals, but do not isolate a shared reward signal that represents universally desirable behaviors. 
To our knowledge, no existing work explicitly disentangles the underlying reward of heterogeneous agents into universally shared competencies and individual preferences. We address this gap with \textbf{General Reward Inference and Disentanglement (GRID)}.

GRID features a novel deep learning architecture that uses an information-theoretic objective to simultaneously capture individual-specific behaviors and the general behaviors common across agents. Concretely, we assume individual reward functions, available via RL environments or IRL, express individual behaviors. Then, we decompose each into a sum of a general reward and an individual-specific reward. The general reward captures a universally shared structure: what every competent agent does regardless of their personal goal. Performing RL with this general reward yields a generalist agent that has internalized the environmental norms of the observed population. Crucially, this agent serves as a strong prior for downstream adaptation as it already knows what to do and what to avoid and can be fine-tuned for any specific task more efficiently than training from scratch. Unlike traditional specialist agents that must learn fundamental physics and safety constraints from scratch for each new objective, our generalist agent already knows how to operate safely in the environment and can be efficiently fine-tuned to specific tasks.

\textbf{Statement of Contributions.} \textbf{(1)} We introduce the GRID architecture and information-theoretic training objective that jointly disentangles general and specific reward structures from a heterogeneous population of demonstrators pursuing different goals. \textbf{(2)} We provide intuition on general and specific reward structures through interpretable visualizations of the learned reward decomposition across synthetic, multi-agent, and robotic domains, confirming semantic alignment with ground-truth reward components. \textbf{(3)} We demonstrate that the GRID generalist outperforms LfD baselines in broadly applicable competencies and downstream fine-tuning to unseen tasks. 

\begin{figure}[t]
  \centering
  \includegraphics[width=\textwidth]{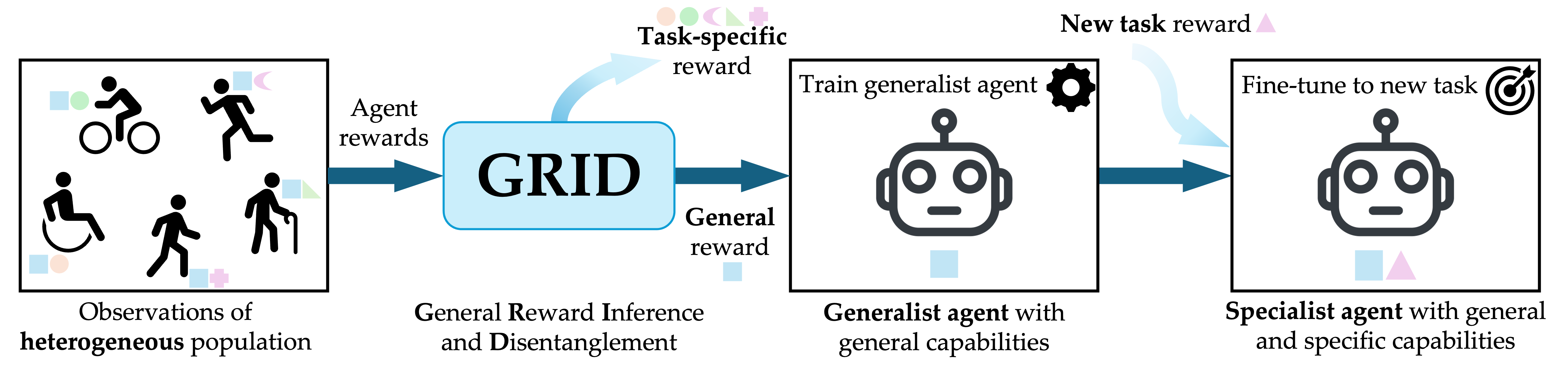}
  \caption{General Reward Inference and Disentanglement (GRID) is a social learning method that extracts universally useful behaviors from a heterogeneous population of demonstrators pursuing different goals. The learned general behavior can be used to train a generalist agent, which can then be fine-tuned for a new task.}
  \label{fig:framework}
\end{figure}

\section{Related Works}
\label{sec:related work}

\textbf{Behavior learning.} Standard LfD methods \citep{TorabiWarnellEtAl2018, RossGordonEtAl2011, AbbeelNg2004, ZiebartMaasEtAl2008, RamachandranAmir2007} and powerful generative approaches \citep{HoErmon2016, FuLuoEtAl2018} share a common assumption that all demonstrations originate from a single, consistent demonstrator. However, in large datasets and in the real world, demonstrations are often heterogeneous, reflecting different skills, preferences, and goals across demonstrators. Later works address heterogeneity through mutual information maximization \citep{LiSongEtAl2017, YuYuEtAl2019, SchrumHedlundEtAl2021, SchrumSumnerEtAl2024, PoddarWanEtAl2024}, while others use CVAEs and variational inference for multi-modal human behavior modeling, common in autonomous driving applications \citep{SchmerlingLeungEtAl2018, IvanovicLeungEtAl2020, RemyChangEtAl2026}. However, all of these approaches treat heterogeneity as a modeling target---the goal is to explain variations between demonstrators. In contrast, our approach explicitly learns the shared commonalities between demonstrators.

\textbf{Reward decomposition.} GRID contributes to a growing interest in structured reward decomposition. Multi-Style Reward Distillation (MSRD) is the closest in spirit to our work \citep{ChenPalejaEtAl2020}. MSRD is an IRL algorithm that learns to perform a single task from heterogeneous demonstrations by decomposing the total reward into a shared task component and an individual strategy component via neural network distillation. Unlike the aforementioned LfD methods, MSRD can explain both the heterogeneity and shared commonalities of various demonstrations. GRID also separates a shared reward from individual rewards, but does not assume all demonstrations are of the same task. In fact, GRID explicitly handles heterogeneous tasks by performing reward decomposition after the IRL procedure instead of during as MSRD does. Furthermore, GRID's approach is grounded in information theory whereas MSRD is not. 
Fast Lifelong Adaptive Inverse Reinforcement learning (FLAIR)  and Reward Decomposition From Suboptimal and Heterogeneous Demonstrations (REPRESENT)  build off of MSRD to better handle various strategies over time \citep{ChenJayanthiEtAl2022} and suboptimal demonstrations \citep{XueChen2025} respectively. 
Other works feature hand-engineered reward decomposition \citep{SeijenFatemiEtAl2017}, redistribute rewards temporally along relevant points in the episode \citep{ArjonaMedinaGillhoferEtAl2019}, or seek interpretable decompositions of rewards \citep{JuozapaitisKoulEtAl2019}. GRID unifies these threads: the decomposition is learned without manual reward engineering, is semantic (general vs. specific), and directly supports both interpretability and downstream specialization.

\textbf{Generalist pretraining and fine-tuning.} The paradigm of pretraining a generalist model and adapting it to specific downstream tasks has become central to modern machine learning and robotics \citep{PalanShevchukEtAl2019}. \citet{JaquesGuEtAl2017} introduces KL-control fine-tuning to prevent catastrophic forgetting of a pretrained prior, a technique later adopted for aligning LLMs with human preferences \citep{OuyangWuEtAl2022}. \citet{FinnAbbeelEtAl2017} learns a parameter initialization for rapid adaptation, and \citet{RakellyZhouEtAl2019} extends this by inferring latent task representations without task labels at test time. In contrast, GRID instantiates this paradigm through reward structure: the generalist is trained on the shared, general reward, and specialization consists of fine-tuning on the specific reward, which encodes individual preferences. This grounds the prior in a semantically meaningful decomposition, making the source of generalization transparent and the adaptation target well-defined.

\section{Preliminaries}


\textbf{Markov Decision Process.}
Given a domain with heterogeneous tasks, the goal is to learn a policy that enacts shared environmental norms from heterogeneous demonstrations of those tasks. We assume $N$ demonstrators act according to their own task $p$. Each demonstrator provides $M$ demonstrations for a total of $N \times M$ demonstrations within the same domain. For each demonstrator's task, we model their environment as a Markov Decision Process, denoted by $\langle \mathcal{S}, \mathcal{A}, R^{(p)}, T, \gamma \rangle$; $\mathcal{S}, \mathcal{A}$ represents the state and action space respectively; $R^{(p)}(s,a)$ is the reward for demonstrator ${p}$ for taking action $a$ in state $s$; $T: \mathcal{S} \times \mathcal{A} \times \mathcal{S}\rightarrow [0,1] $ is the transition function that yields the probability of entering the next state $s'$ from the current state $s$ by applying action $a$; and the total discounted reward over trajectory $\tau=\langle s_0,a_0, ..., s_T,a_T\rangle$ is $R^{(p)}(\tau)=\sum_{t=0}^T\gamma^t R^{(p)}(s_t,a_t)$ where $\gamma \in [0,1)$ is the discount factor. The goal of reinforcement learning is to learn an optimal policy $\pi^*: \mathcal{S} \rightarrow \mathcal{A}$ that maximizes the expected future reward: $\pi^*=\arg\max_{\pi}\mathbb{E}_{\tau\sim\pi}[R(\tau)]$. \citet{Ziebart2010} introduces the principle of maximum entropy that appends an entropy bonus to the standard cost function to incentivize exploration: $\pi^*=\arg\max_{\pi}\mathbb{E}_{\tau\sim\pi}[\sum_{t=0}^T R(s_t, a_t) + \alpha H(\pi (\cdot|s_t))]$.

\textbf{Inverse Reinforcement Learning.} 
Standard IRL operates within an MDP$\backslash R$ and infers a reward function that explains a set of demonstrated trajectories $\mathcal{D}=\{\tau_1, .., \tau_M\}$. Maximum entropy IRL infers a reward function that exponentially favors the demonstrated trajectories according to the probability distribution: $p_\theta(\tau) \propto \exp(\sum_{t=0}^T \gamma^t R_\theta (s_t,a_t))$. Then, the IRL procedure seeks to maximize the likelihood: $\max_\theta \mathbb{E}_{\tau \sim \mathcal{D} } [\log p_\theta (\tau)]$. In our problem set up, the dataset provided is $\mathcal{D}=\{\mathcal{D}^1, ..., \mathcal{D}^N\}$ where $\mathcal{D}^{p}=\{\tau^p_1,...\tau^p_M\} \forall p \in \{1, ..., N\}$. We can perform IRL on each $\mathcal{D}^p$ to yield $R^{(p)}$ for each task.

\textbf{Information Bottleneck.} 
Given an input $X$ and output $Y$, the goal of an information bottleneck (IB) is to learn an encoding $Z$ from $X$ that is maximally expressive about $Y$ while being maximally compressive about $X$ \citep{TishbyPereiraEtAl1999}. The IB principle \eqref{eq:ib loss} describes expressiveness and compressibility with mutual information, $I$, and describes their tradeoff with a positive hyperparameter $\beta$. However, IB is difficult to implement in practice because computing the mutual information is challenging. To address this, \citet{AlemiFischerEtAl2019} develop a deep variational information bottleneck (VIB) using variational inference to bound the mutual information. These bounds are described by a variational decoder $q(y|z)$, the entropy of $Y$, $H(Y)$, and the KL divergence between the encoder distribution $p(Z|X)$, and a prior distribution $r(z)$ (see \eqref{eq:lower} and \eqref{eq:upper}). 
\begin{equation}
    \label{eq:ib loss}
    \mathcal{L}_{IB} = I (Z;Y)- \beta I (Z;X)
\end{equation}
\begin{equation}
    \label{eq:lower}
    I (Z;Y) \geq \mathbb{E}_{y,z \sim p(y,z)} [\log q (y|z)] + H(Y)
\end{equation}
\begin{equation}
    \label{eq:upper}
    I(Z; X) \leq \mathbb{E}_{x \sim p(x)} [D_{KL}(p(z|x) || r(z))]
\end{equation}

\section{General Reward Inference and Disentanglement}
\begin{figure}[t]
    \centering
    \includegraphics[width=0.91\textwidth]{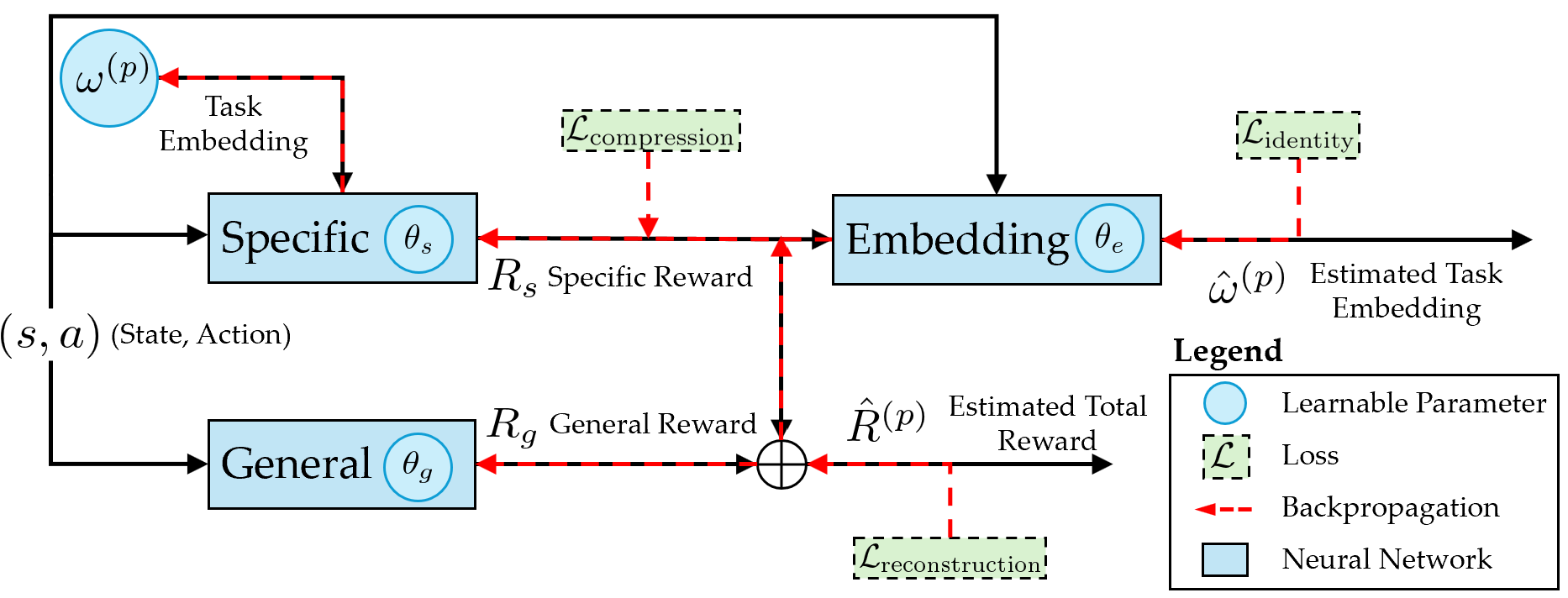}
    \caption{The model architecture of GRID. The summed outputs of the general and specific reward models yield the total reward. $\mathcal{L}_{\text{compression}}$ and $\mathcal{L}_{\text{identity}}$ form the information bottleneck while $\mathcal{L}_{\text{reconstruction}}$ helps predict the total reward. Losses are jointly optimized.} 
    \label{fig:model}
\end{figure}

We present General Reward Inference and Disentanglement (GRID), a novel IRL framework (Fig. \ref{fig:framework}) and its architecture (Fig. \ref{fig:model}) that can be used to train a generalist agent by extracting the general reward signal from a heterogeneous population of demonstrators performing different tasks in the environment. GRID uses an information-theoretic technique called the variational information bottleneck to disentangle $R^{(p)}$ into $R_g$ and $R_s^{(p)}$.

\subsection{Problem Setup}
We consider multiple tasks with different reward functions in the same domain. We assume that provided demonstrations are optimal with respect to the individual tasks and representative of environmental norms. Behaviors that are not present in these demonstrations are then not representative of environmental norms. After we acquire a reward signal $R^{(p)}$ for each demonstrated task $p$, we seek to disentangle the general reward $R_g$ from the specific reward $R_s^{(p)}$.
We call the shared component the general reward because it describes how any agent should \textit{generally} act in the domain, and we call the individual component the \textit{specific} reward because it describes the agent's specific task. Thus, the reward corresponding to task $p$ is $R^{(p)} = R_g + R_s^{(p)}$. By definition, the specific $R^{(p)}_s$ is defined by task $p$, meaning the inputs are the task identifier and the states and actions. Additionally, we seek to learn the rewards such that $R_s^{(p)}$ contains the minimum amount of shared information across all tasks. In other words, $R_s^{(p)}$ should be sparse so $R_g$ can encompass all of the domain information relevant to environmental norms and general capabilities. $R_g$  is completely independent of what task is performed and is purely a function of states and actions. We parameterize $R_s^{(p)}$ and $R_g$ as neural network parameters $\theta_s$ and $\theta_g$. Lastly, we embed each task identity $p$ into a learnable embedding vector $\omega^{(p)}\in \mathbb{R}^d$ for greater expressivity. The overall function decomposition becomes:
\begin{equation}
    \label{eq:reward-decomp}
    R^{(p)}=R_g(s,a; \theta_g) + R_s^{(p)} (\omega^{(p)},s,a; \theta_s)
\end{equation}
\subsection{GRID Architecture and Target Requirements}
\label{sec:grid}
Following \eqref{eq:reward-decomp}, the GRID architecture (\figref{fig:model}) features two neural network blocks parameterized by $\theta_g$ and $\theta_s$, that represent $R_g$ and $R_s^{(p)}$ respectively, whose outputs are summed to reconstruct the total reward: $\hat{R}^{(p)}(\omega^{(p)},s,a)=R_g(s,a)+R_s^{(p)}(\omega^{(p)},s,a)$. The architecture is trained to satisfy three requirements:
\textbf{(R1)} The specific reward $R_s^{(p)}(s,a,\omega^{(p)})$ captures behaviors relevant to corresponding agent $p$. 
\textbf{(R2)} The general reward captures behaviors that are meaningful to \emph{all} agents. In other words, the general reward should primarily explain the corresponding behavior of each task, leaving the specific reward to be emphasized only when necessary.
\textbf{(R3)} It must accurately reproduce the original reward $R^{(p)}$. That is, $R^{(p)} \approx \hat{R}^{(p)}$.
We address each of these requirements by jointly optimizing the information bottleneck and reconstruction loss described below.

\textbf{Information Bottleneck.} We utilize the IB formulation to address \textbf{(R1)} and \textbf{(R2)}. The specific reward should be maximally expressive of the task at hand, $p$, while maximally compressive so that $R_g$ fully represents environmental norms. Succinctly, we let the input $X=(\omega^{(p)}, s, a)$ be the embedding and the state and action; the encoding $Z=(R_s^{(p)})$ be the specific reward; and the target $Y=(\omega^{(p)})$ be the identifier. While standard VIB defines the decoder purely on the bottleneck $q(Y|Z)$, a scalar reward value of $R_s^{(p)}$ cannot reconstruct a continuous task embedding without environment context. Therefore, we implement a \textit{conditional} formulation of the lower bound, where the auxiliary embedding network $q$ is conditioned on states and actions. Below outlines the resulting bounds, where $q (\omega^{(p)} | R_s^{(p)}, s, a; \theta_e)$ is the auxiliary embedding network parameterized by $\theta_e$ and $r(z)$ is a chosen prior representative of the specific reward.
\begin{equation}
\label{eq:lower_grid}
I (R_s;\omega^{(p)}|s,a) \geq \mathbb{E}_{s,a,p \sim D} [\log q (\omega^{(p)}|R_s^{(p)}, s,a)] + H(\omega^{(p)} \mid s, a)
\end{equation}
\begin{equation}
\label{eq:upper_grid}
I(R_s; \omega^{(p)}, s,a) \leq \mathbb{E}_{s,a,p \sim D}[D_{KL}(p(R_s^{(p)}|\omega^{(p)},s,a) || r(z))]
\end{equation}
Maximizing \eqref{eq:lower_grid} is equivalent to minimizing \eqref{loss:identity} below and backpropagating the gradients into the original $\omega^{(p)}$ \citep{SchrumHedlundEtAl2021, SchrumSumnerEtAl2024}. To do so, we sample $\omega^{(p)}\sim\mathcal{N}(0,I_d)$ before training. Then, during training, $q$ attempts to reproduce $\omega^{(p)}$, incentivizing its prediction $\hat{\omega}^{(p)}$ and $\omega^{(p)}$ to converge to each other's values.
\begin{equation}
    \label{loss:identity}
    \mathcal{L}_{\text{identity}} (\theta_s, \theta_q, \omega^{(p)}) = \mathbb{E}_{s,a,p \sim D} \biggr[\frac{1}{d}\sum_{i=1}^d(\hat{\omega}_i^{(p)}-\omega_i^{(p)})^2 \biggr ]
\end{equation}
Choosing $r(z)$ to be the zero-centered Laplace distribution simplifies minimizing \eqref{eq:upper_grid} into minimizing the $L1$ loss of $R_s^{(p)}$. Alternatively, choosing $r(z)$ to be the zero-meaned Gaussian turns it into minimizing the $L2$ loss of $R_s^{(p)}$. We present the $L1$ variant below.
\begin{equation}
    \mathcal{L}_{\text{compression}}(\theta_s) =\mathbb{E}_{s,a,p \sim D} [|R_s^{(p)}(\omega^{(p)},s,a)|]
    \label{loss: emph}
\end{equation}

In summary, the VIB loss ensures the general reward captures as much information as possible, but anything that is not shared among agents is compressed into the agent-specific rewards.

\textbf{Reconstruction Loss.} To address \textbf{(R3)}, we minimize the following squared error loss between the predicted total reward $\hat{R}^{(p)}$ and the provided reward $R^{(p)}$ for faithful decomposition of the general and specific components to the observed reward signal.
\begin{equation}
    \mathcal{L}_{\text{reconstruction}}(\theta_g, \theta_s, \theta_q, \omega^{(p)}) = \mathbb{E}_{s,a,p \sim D} [(\hat{R}^{(p)}(\omega^{(p)},s,a)-R^{(p)})^2]
    \label{loss:recon}
\end{equation}
Altogether, we combine the IB losses and reconstruction loss to jointly optimize the network parameters: $\mathcal{L}_{\text{total}}=  \mathcal{L}_{\text{reconstruction}} + \lambda_1 \mathcal{L}_{\text{identity}} + \lambda_2 \mathcal{L}_{\text{compression}}$, where $\lambda_1, \lambda_2 > 0$.
\begin{figure}[t!]
    \centering
    \includegraphics[width=\linewidth]{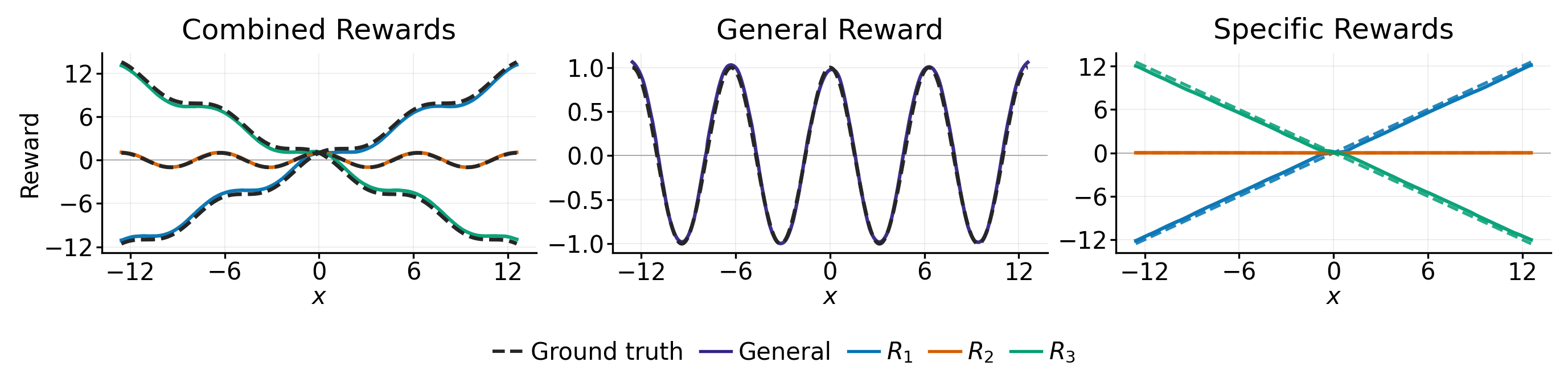}
    \caption{Basis Function Decomposition. Left shows the total reward function. Middle shows the general reward learned by GRID overlapping the ground-truth $\cos(x)$ reward. Right shows the specific rewards learned by GRID overlapping the ground-truth linear $\{+1x, +0x, -1x\}$ rewards. As shown, GRID can automatically and precisely disentangle the sinusoidal and linear behaviors.}
    \label{fig:basis-func}
\end{figure}
\subsection{Generalist and Specialist Training}

\textbf{Generalist Training.} 
We train a generalist agent by applying any RL algorithm to the learned general reward component $R_g$. Following the principle of maximum entropy, RL agents explore the environment during training, potentially leading the agent to states that are relatively out of distribution (OOD) compared to the demonstrations. In this region, GRID's general reward may not extrapolate well, so we develop a mixture of rewards for explored states in the RL process. The mixture contains a $[-1, 1]$ scaled general reward $\Tilde{r}(s,a)$ and OOD reward $r_{ood}$ that are weighted by $w(s)$ and $1-w(s)$: $r(s,a)=w(s)\Tilde{r}(s,a) + (1-w(s))r_{ood}$. For discrete state space domains, $w(s)=1$ for encountered states in the demonstrations and $w(s)=0, r_{ood}=-1$ for never encountered states. For continuous state spaces, we use a $ k$-nearest-neighbors estimate of demonstrated state support. $w(s)$ is close to 1 for states near demonstrated states and approaches 0 as the state moves away from demonstrated states. We set $r_{ood} \leq -1$, so the agent moves toward in-distribution states that are accurately characterized by GRID. This is similar to \citet{ReddyDraganEtAl2020}'s idea: states in the demonstrations are rewarded +1 while all other states are rewarded 0. 

\textbf{Specialist Training.} Once the generalist is trained, we fine-tune the generalist policy to create a specialist agent. We use \citet{JaquesGuEtAl2017}'s KL-control method to fine-tune the generalist policy $\pi_g$ for a new task defined by a new specialist reward function $ R_{FT} (s_t, a_t)$. The fine-tuning objective maximizes the reward while penalizing divergence from the generalist prior $\pi_g$: $\mathcal{L}(\pi_{FT})=\mathbb{E}_\pi \left[\sum_t R_{FT}(s_t,a_t) + \beta \big( \log\pi_g(a_t \mid s_t) - \log\pi_{FT}(a_t\mid s_t) \big) \right]$. The KL penalty prevents catastrophic forgetting of the general competencies encoded in $\pi_g$, while allowing the specialist policy to deviate toward agent-specific preferences. This mirrors the same use of fine-tuning in language models \citep{OuyangWuEtAl2022}, but grounds the prior in a semantically decomposed reward signal rather than a pretraining corpus. 
\begin{figure}[t]
    \centering
    \includegraphics[width=\textwidth]{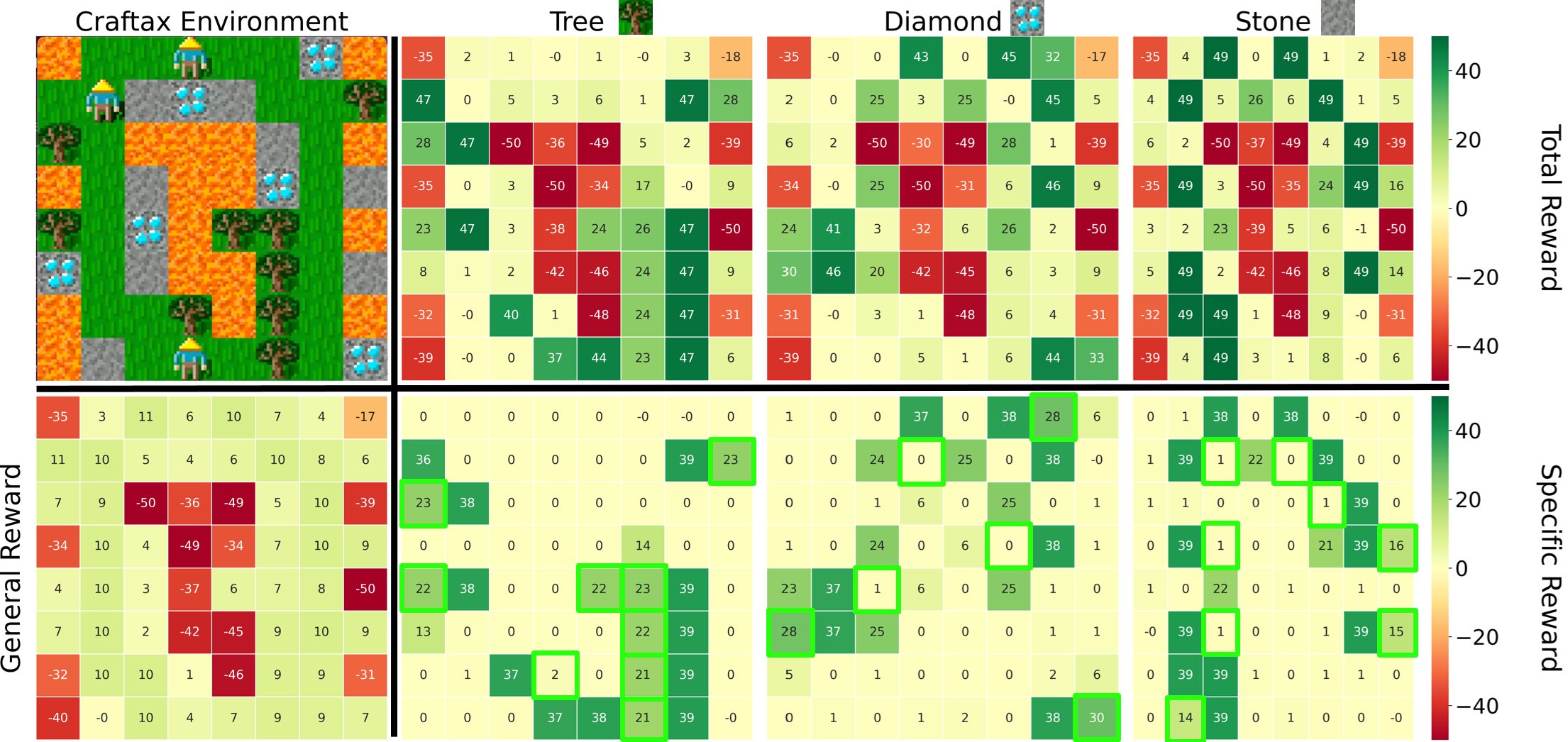}
    \caption{Craftax Domain. We display $\max_a R(s,a)$ for total, specific, and general rewards learned by GRID over the state space for each agent. Penalties are issued for being on the lava tile while collecting the specified item (tree, diamond, or stone) is rewarded. Items can be collected by either being on the actual tile or its adjacent tiles. Tiles with the specific item are boxed in neon green for visual aid.
    The general reward incentivizes avoiding tiles with lava, while the specific rewards incentivize mining for trees, diamonds and stone on tiles close to those resources, showing GRID effectively learns shared safety norms while preserving individual tasks.
    }
    \label{fig:craftax_result}
\end{figure}
\section{Experiments}

\subsection{Domains}
\textbf{Basis Function Decomposition.} A simple synthetic continuous domain with known ground truth designed to provide clear intuition for reward disentanglement (see \figref{fig:basis-func}). We define three total reward functions: $R_1(x) = \cos(x) + x$, $R_2(x) = \cos(x)$, and $R_3(x) = \cos(x) - x$ where the general reward is $\cos(x)$ and specific rewards are $\{+1x, +0x, -1x\}$. GRID is trained to reconstruct these rewards over the range $[-4\pi, 4\pi]$. We additionally inject noise into ground truth values to assess robustness. We report the mean squared error (MSE) for the general reward. 

\textbf{Craftax.} An $8 \times 8$ multi-agent discrete world, representing a simplified version of the Minecraft game \citep{YeJaques2024}. Three agents each pursue a unique objective: collect wood, stone, or diamond, and all share a penalty for stepping on lava. This domain tests GRID's ability to disentangle a shared safety constraint from different individual goals in a discrete state-action space. We visualize the reward distribution across the state space for general and specific rewards. 

\textbf{Highway-Env.} A continuous state and discrete action space autonomous driving simulator, modified into a three-agent setting with four lanes and additional moving vehicles \citep{Leurent2018}. All agents share a general reward for maintaining high speed and avoiding crashes. Each agent additionally holds an easily definable preference: Agent $i$ prefers Lane $i$, for $i=0,1,2$. Lane 3 is deliberately withheld from training to evaluate downstream specialization to an unseen preference. We perform AIRL for each agent to acquire rewards.

\textbf{MuJoCo.} Two continuous state and action domains: pendulum and hopper \citep{gymnasium_robotics2023github}. Traditionally, these domains have a single reward for a single task. To create multiple diverse tasks, we follow \citet{ChenPalejaEtAl2020}'s example of linearly combining the main environment reward with the objective from Diversity is All You Need (DIAYN) \citep{EysenbachGuptaEtAl2019} which learns a discriminator $q_\phi (z|s)$ that distinguishes various behaviors according to the agent's state as part of the reward, thereby promoting unique behaviors. 
\begin{equation}
    r_z(s,a)=\log q_\phi (z | s) - \log p(z)
\end{equation}
To develop broader heterogeneous behavior, we heavily emphasize the DIAYN reward when combined with the main environment reward. Overall, we generate 19 tasks for pendulum and 13 tasks for hopper. We perform AIRL on each task to acquire rewards.

\subsection{Baselines and Comparison Metrics.}
For comparison methods, we consider standard LfD approaches, behavior cloning (BC) \citep{TorabiWarnellEtAl2018} and Adversarial Inverse Reinforcement Learning (AIRL) \citep{FuLuoEtAl2018}, which aim to learn a policy from the heterogeneous demonstrations. We also compare against MSRD \citep{ChenPalejaEtAl2020}, FLAIR\footnote{For FLAIR specifically, we do not learn policies sequentially but in parallel since GRID is not a life long learning algorithm. } \citep{ChenJayanthiEtAl2022}, and REPRESENT \citep{XueChen2025} which are most similar to GRID (see \secref{sec:related work}) as they too learn a decomposed reward from heterogeneous demonstrations but of the \textit{same} task. 
For the highway-env and MuJoCo environments, we compare episode returns as a general notion of success. For highway-env specifically, we report lane metrics: lane entropy and lane 3 steps. Entropy measures the generalist's bias. Having a strong bias toward certain states (low entropy) is an indicative heuristic of having more difficulty fine tuning to a new preference (lane 3). We evaluate specialist capabilities by training the generalist policy with KL control. In highway-env, $R_{FT}$ is the lane 3 reward while in MuJoCo, $R_{FT}$ is the DIAYN reward. The total of the true domain rewards and fine tune rewards are used to evaluate the specialist policy. 
\begin{figure}[t]
    \centering
    \begin{subfigure}{0.18\textwidth}
        \centering
            \begin{subfigure}{\textwidth}
            \centering
            \includegraphics[width=\textwidth]{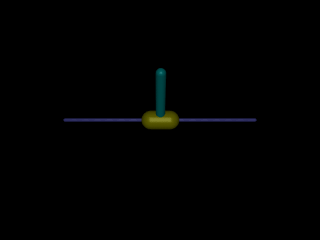}
        \end{subfigure}
        \begin{subfigure}{\textwidth}
            \centering
            \includegraphics[width=\textwidth]{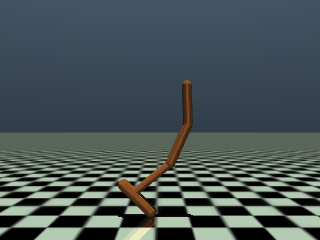}
        \end{subfigure}
    \end{subfigure}
    \begin{subfigure}{0.8\textwidth}
        \centering
        \includegraphics[width=\textwidth]{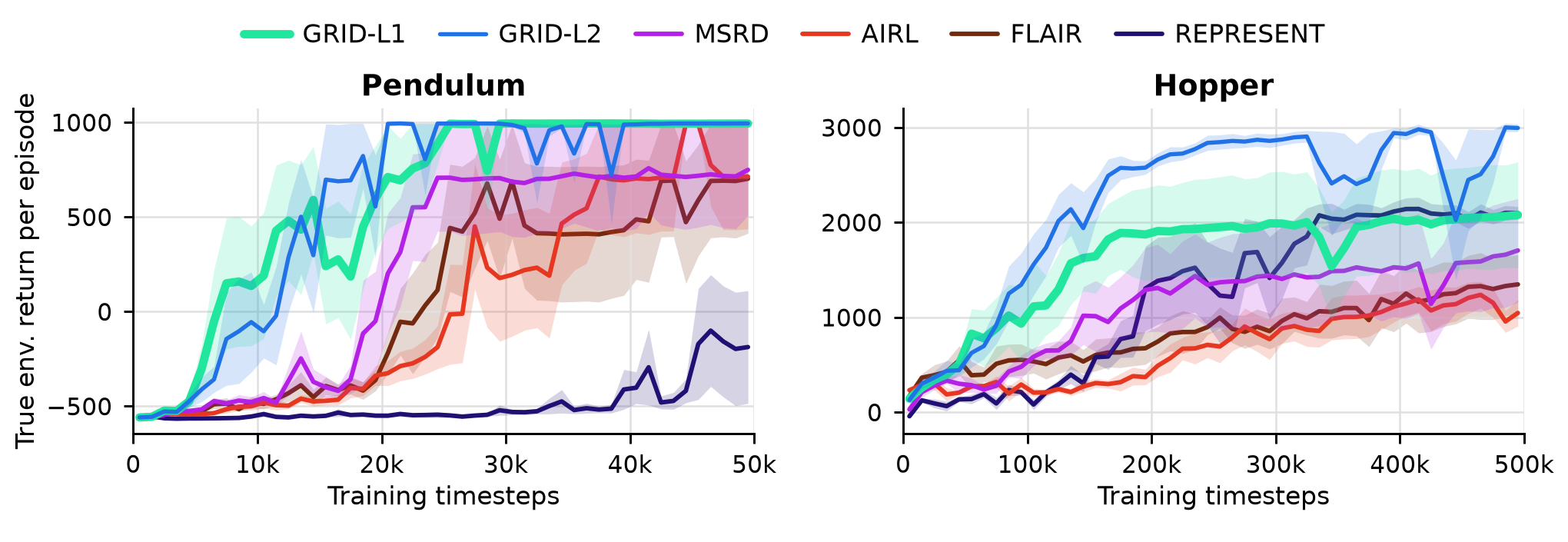}
    \end{subfigure}
  \caption{Left figure depicts the pendulum and hopper domains. Middle and right plots show SAC returns throughout training for both respectively. SAC is trained on GRID's general reward, MSRD's task reward and AIRL's learned reward. GRID-L2 variant achieves the best performance. }
  \label{fig:sac}
\end{figure}

\begin{table}[h!]
  \caption{Generalist and specialist comparisons in the Highway-Env. See Tables \ref{tab:gen} and \ref{tab:spec} in the Appendix for the full lane distributions. Means and standard errors are shown.}
  \centering
  \setlength{\tabcolsep}{4pt}
  \resizebox{\textwidth}{!}{%
  \begin{tabular}{l cc cccc}
    \toprule
    & 
    \multicolumn{2}{c}{\textbf{Generalist}} & 
    \multicolumn{4}{c}{\textbf{Specialist}} \\
    \cmidrule(lr){2-3} \cmidrule(lr){4-7}
    \textbf{Method} & 
    \makecell{\textbf{General} \\ \textbf{Reward ($\uparrow$)}} & 
    \textbf{Entropy ($\uparrow$)} &
    \makecell{\textbf{General} \\ \textbf{Reward ($\uparrow$)}} & 
    \makecell{\textbf{Specialist} \\ \textbf{Reward ($\uparrow$)}} & 
    \makecell{\textbf{Total} \\ \textbf{Reward ($\uparrow$)}} &
    \makecell{\textbf{Lane 3} \\ \textbf{Steps ($\uparrow$)}} \\
    \midrule
    BC \citep{TorabiWarnellEtAl2018} & $7.12 \pm 0.24$ & $\mathbf{1.84 \pm 0.01}$ & $3.94 \pm 0.29$ & $\mathbf{3.44 \pm 0.06}$ & $7.39 \pm 0.26$ & $\mathbf{34.5 \pm 0.6}$ \\
    AIRL \citep{FuLuoEtAl2018} & $12.62 \pm 0.43$ & $1.62 \pm 0.09$ & $6.92 \pm 0.46$ & $3.30 \pm 0.06$ & $10.23 \pm 0.40$ & $33.1 \pm 0.6$ \\
    MSRD \citep{ChenPalejaEtAl2020} & $12.43 \pm 0.28$ & $1.70 \pm 0.11$ & $6.56 \pm 0.29$ & $3.36 \pm 0.09$ & $9.91 \pm 0.23$ & $33.7 \pm 0.9$ \\
    FLAIR \citep{ChenJayanthiEtAl2022}& $12.38 \pm 0.07$ & $1.82 \pm 0.06$ & $7.12 \pm 0.34$ & $3.26 \pm 0.11$ & $10.37 \pm 0.24$ & $31.4 \pm 0.7$ \\
    REPRESENT \citep{XueChen2025} & $12.38 \pm 0.29$ & $1.77 \pm 0.04$ & $8.00 \pm 0.91$ & $3.05 \pm 0.13$ & $11.05 \pm 0.89$ & $30.6 \pm 1.3$ \\
    GRID-L1 (ours) & $\mathbf{13.41 \pm 0.18}$ & $1.77 \pm 0.04$ & $8.40 \pm 0.23$ & $2.97 \pm 0.10$ & $11.37 \pm 0.24$ & $29.8 \pm 1.0$ \\
    GRID-L2 (ours) & $12.68 \pm 0.36$ & $1.74 \pm 0.07$ & $\mathbf{8.60 \pm 0.53}$ & $2.83 \pm 0.03$ & $\mathbf{11.43 \pm 0.51}$ & $28.5 \pm 0.3$ \\
    \bottomrule
  \end{tabular}
  }
  \label{tab:combined_results}
\end{table}
\section{Results and Discussion}
\textbf{Basis Function Decomposition.} The learned general and specific functions using GRID are shown in \figref{fig:basis-func}. We can see that the disentangled $\cos(x)$ behavior in the general reward and the linear behaviors in the specific rewards are well-recovered by GRID. We note that while GRID performs supervised learning using the ground truth total reward signal, the general and specific rewards are learned in an unsupervised manner and that the $L1$ variant performs better than $L2$ in this domain. With no noise injection on the ground truth, the MSE of the general reward is $0.0018$. Even with a $\mathcal{N}(0,1)$ Gaussian noise injection, the corresponding error is $0.0298$. Additional sets of basis functions are in the \appref{app:basis-func}.

\textbf{Craftax.} We show clear agent behavior disentanglement in \figref{fig:craftax_result}. Since the gridworld has discrete states and actions, we report $\max_a R(s,a)$ in each grid cell. The general reward correctly identifies lava locations as shared negative penalties across all agents, while item-gathering preferences (tree wood, diamond, and stone) are cleanly partitioned into the agent-specific reward maps with no observable overlap. We note that the specific reward does not necessarily coincide exactly with the exact item locations, because items can be collected (i.e., reward gained) if the agent is adjacent to those items. Thus, the maximum reward for any tile adjacent to an item can be gained by taking the collect item action. In contrast, the agent takes a penalty regardless of the action taken if it is on lava.

\textbf{Highway-Env.} We evaluate the entire framework on this domain with interpretable rewards to demonstrate GRID's effectiveness. \Tabref{tab:combined_results} shows the full results. We collect demonstrations from agents $0-2$, perform AIRL, apply GRID to learned rewards, and train generalist and specialist (prefer lane 3) policies using DQN \citep{MnihKavukcuogluEtAl2013}. While the BC prior has the highest lane entropy and occupies lane 3 the most in the fine-tuning stage, it offers the worst overall performance. The BC policies learn to stay in lane 3 but slow down when they are behind another vehicle and do not move around it to maintain high speeds. In contrast, while GRID spends the least amount of steps in lane 3 ($\approx 30$ of 40 steps), the GRID specialist is still able to maneuver around cars in front of it. In fact, the agent spends less time in lane 3 because it moves around slower vehicles in that lane. AIRL and MSRD sacrifice their agility around other vehicles to stay in lane 3, resulting in lower general rewards after fine-tuning. FLAIR and REPRESENT outperform those two methods but not GRID.

\textbf{MuJoCo.} We evaluate the GRID framework on significantly more tasks. We use AIRL to learn the DIAYN tasks (19 pendulum and 13 hopper), apply GRID to learned rewards, and train the generalists and specialist policies using SAC \citep{HaarjonaZhouEtAl2018-key}. Here, the rewards are less interpretable, so we only report rewards.
\Figref{fig:sac} displays the environment ground truth learning curves for GRID, MSRD, and AIRL. AIRL has no ability to disentangle a shared reward from a specific reward, so its reward function is forced to explain multiple heterogeneous demonstrations. Thus, it expectantly performs subpar to GRID and MSRD. Both GRID $L1$ and $L2$ variants surpass MSRD's learning ability. We attribute this to GRID's separate staged process of acquiring rewards first and then disentangling rewards, where there is no competition between reward learning and disentangling. In contrast, MSRD, FLAIR and REPRESENT, which are built on multiple AIRL models, have to jointly adversarially learn and distill reward information. We perform one fine tuning task in pendulum and two tasks in hopper. Similar to our previous results, GRID generally performs on par with or better than all baselines as shown in \tabref{tab:finetune-mujoco}. We note in Hopper New Task 2, GRID-L2's best seed average return is around double that of MSRD and FLAIR, demonstrating GRID's ability to learn a strong generalist prior for fine tuning.

\begin{table}[t]
\centering
\caption{Generalist and fine-tuning performance on MuJoCo environment. We fine tune generalist priors to new tasks using SAC and KL-control. We fine tune one new task in pendulum for 30k steps and two new tasks in hopper for 100k steps. The best seed-averaged returns across training and peak returns are reported across 5 seeds. GRID tends to perform similarly if not better than baselines.}
\label{tab:finetune-mujoco}
\resizebox{\textwidth}{!}{%
\begin{tabular}{@{}l c c cccccc@{}}
\toprule
\multirow{2}{*}{\textbf{Method}} & 
\multirow{2}{*}{\makecell{\textbf{Pendulum} \\ \textbf{General@50k Steps}}} & 
\multirow{2}{*}{\makecell{\textbf{Hopper} \\ \textbf{General@500k Steps}}} & 
\multicolumn{2}{c}{\textbf{Pendulum New Task}} & \multicolumn{2}{c}{\textbf{Hopper New Task 1}} & \multicolumn{2}{c}{\textbf{Hopper New Task 2}} \\
\cmidrule(lr){4-5} \cmidrule(lr){6-7} \cmidrule(l){8-9}
 & & & \textbf{Seed-Avg} & \textbf{Peak} & \textbf{Seed-Avg} & \textbf{Peak} & \textbf{Seed-Avg} & \textbf{Peak} \\
\midrule
BC \citep{TorabiWarnellEtAl2018} & -- & -- & 1399.0 $\pm$ 14.9 & 1427.9 & 590.6 $\pm$ 167.6 & 936.3  & 1433.6 $\pm$ 676.8 & 3062.3 \\
AIRL \citep{FuLuoEtAl2018} & 713.2 $\pm$ 278.8 & 1000.8 $\pm$ 125.8 & 1402.7 $\pm$ 11.0 & 1425.6 & 748.1 $\pm$ 81.1  & 964.1  & 266.3 $\pm$ 306.2  & 1526.6 \\
MSRD \citep{ChenPalejaEtAl2020} & 731.6 $\pm$ 258.7 & 1682.1 $\pm$ 551.5 & 1398.4 $\pm$ 15.6 & \textbf{1452.2} & 646.1 $\pm$ 175.4 & 987.3  & 1547.5 $\pm$ 803.7 & 3709.3 \\
FLAIR \citep{ChenJayanthiEtAl2022} & $696.1 \pm 295.3$ & $1338.2 \pm 312.0$ & $1399.5 \pm 9.8$ & $1442.8$ & \textbf{801.2 $\pm$ 19.9} & $866.7$ & $1321.9 \pm 524.4$ & $3014.3$\\
REPRESENT \citep{XueChen2025} & $-192.5 \pm 298.7$ & $2099.1 \pm 71.2$ & $1372.1 \pm 11.9$ & $1447.4$ & $766.8 \pm 15.0$ & $812.1$ & $1821.4 \pm 517.7$ & $3101.2$\\
GRID-L1 (ours) & 993.6 $\pm$ 0.9 & 2072.1 $\pm$ 548.7 & \textbf{1405.5 $\pm$ 14.2} & 1440.5 & 680.4 $\pm$ 355.5 & \textbf{2088.9} & 2006.3 $\pm$ 624.8 & 3788.8 \\
GRID-L2 (ours) & \textbf{994.1 $\pm$ 0.5} & \textbf{2997.6 $\pm$ 40.9} & 1402.9 $\pm$ 16.9 & 1449.6 & 678.1 $\pm$ 113.8 & 1001.2 & \textbf{2959.8 $\pm$ 518.6} & \textbf{5228.3} \\
\bottomrule
\end{tabular}
}
\end{table}

\section{Conclusion, Limitations, and Future Work}
\label{sec:conclusion}
We introduce General Reward Inference and Disentanglement (GRID), a method for social learning that determines which behaviors are worth acquiring from a heterogeneous population of agents. GRID successfully disentangles reward signals into a general component that captures useful universal behaviors and a specific component that encodes individual preferences. We demonstrate that performing RL on the extracted general reward yields a generalist agent that internalizes fundamental environmental norms and that the generalist serves as a prior prime for fine-tuning. 
With the exception of highway-env, the domains tested feature agents that typically act independently from one another. Studying more interactive multi-agent settings is an important direction for future work, although learning rewards in these environments may be more challenging. 
Since GRID requires task rewards, its disentanglement ability is directly related to the quality of the rewards. In this work, GRID receives rewards from AIRL and therefore inherits AIRL's assumption that demonstrations are optimal. Extending GRID to accommodate suboptimal demonstrations, for example, by drawing on the approaches of \citet{ChenPaleja2020} and \citet{XueChen2025}, would be an interesting direction for future research.
More broadly, in the context of social learning, enabling a generalist agent to learn from surrounding agents that are interactive, heterogeneous, and potentially suboptimal presents an exciting opportunity. 
Additionally, this paper focuses on the general reward and its applications, leaving the specific rewards and task embeddings largely unexplored. Developing methods to use these embeddings to characterize demonstrators could enable further personalization, such as in assistive-AI settings, as well as targeted fine-tuning.

\newpage
\subsection*{AI use statement}


In this work, we used generative AI tools to help run RL algorithms from stable-baselines3 to collect data, implement baselines, provide implementation suggestions for handling OOD states, and aid in the IB loss derivation. We have not used generative AI tools for the ideation and implementation of our GRID framework and architecture, cleaning and reformatting datasets, supporting qualitative and thematic data analysis, or interpreting results and formulating mathematical claims, providing critical ingredients for proving mathematical claims, assisting in the writing of proofs, proposing or refining hypotheses, and assisting with translation are not applicable to this work.

Additionally, we used generative AI tools for formatting tables and figures, editing writing, and finding related works. We have reviewed all AI-assisted work. For example, RL results were confirmed, visualized, and cross referenced with results from other works and two authors went over the loss derivation. We take responsibility for the final content of this work,
including text, claims or artifacts produced with the aid of generative AI.


\subsection*{Ethics statement}
Our work does not raise ethical concerns.


\subsection*{Reproducibility statement}
The implementation of the GRID architecture and training scripts are provided for the basis function decomposition domain.


\subsubsection*{Author Contributions}
If you'd like to, you may include  a section for author contributions as is done
in many journals. This is optional and at the discretion of the authors.

\subsubsection*{Acknowledgments}
Use unnumbered third level headings for the acknowledgments. All
acknowledgments, including those to funding agencies, go at the end of the paper.

\bibliography{main, ctrl_papers}

@Preamble{"\newcommand{\noopsort}[1]{} " #
"\newcommand{\printfirst}[2]{#1} " #
"\newcommand{\singleletter}[1]{#1} " #
"\newcommand{\switchargs}[2]{#2#1} "}

@String{jrn_IEEE_RAL                   = {{IEEE Robotics and Automation Letters}}}

@String{jrn_IEEE_TR                    = {{IEEE Transactions on Robotics}}}

@String{proc_AAAI_AAAI                 = {{Proc.\ AAAI Conf.\ on Artificial Intelligence}}}

@String{proc_AAAI_FS                   = {{AAAI Fall Symposium}}}

@String{proc_AAAI_IJCAI                = {{Int.\ Joint Conf.\ on Artificial Intelligence}}}

@String{proc_AISTATS                   = {{AI \& Statistics}}}

@String{proc_CoRL                      = {{Conf.\ on Robot Learning}}}

@String{proc_ICLR                      = {{Int.\ Conf.\ on Learning Representations}}}

@String{proc_ICML                      = {{Int.\ Conf.\ on Machine Learning}}}

@String{proc_IEEE_ICRA                 = {{Proc.\ IEEE Conf.\ on Robotics and Automation}}}

@String{proc_NIPS                      = {{Conf.\ on Neural Information Processing Systems}}}

@String{proc_RSS                       = {{Robotics: Science and Systems}}}

@String{pub_MIT                        = {{MIT Press}}}

@InProceedings{SchmerlingLeungEtAl2018,
  author    = {Schmerling, E. and Leung, K. and Vollprecht, W. and Pavone, M.},
  booktitle = proc_IEEE_ICRA,
  title     = {{Multimodal Probabilistic Model-Based Planning for Human-Robot Interaction}},
  year      = {2018},
  arxiv     = {1710.09483},
  category  = {interaction},
  img       = {SchmerlingLeungEtAl2018.png},
  selected  = {true},
}

@Article{IvanovicLeungEtAl2020,
  author   = {Ivanovic, B.* and Leung, K.* and Schmerling, E. and Pavone, M.},
  journal  = jrn_IEEE_RAL,
  number   = {2},
  pages    = {295--302},
  title    = {{Multimodal Deep Generative Models for Trajectory Prediction: A Conditional Variational Autoencoder Approach}},
  volume   = {6},
  arxiv    = {2008.03880},
  category = {interaction},
  img      = {IvanovicLeungEtAl2020.jpg},
  selected = {true},
  year     = {2021},
}

@Article{RemyChangEtAl2026,
  author   = {Remy, I. and Chang, C. and Leung, K.},
  journal  = {{{Available at }\url{https://arxiv.org/abs/2604.13128}}},
  title    = {{Learning Probabilistic Responsibility Allocations for Multi-Agent Interactions}},
  year     = {2026},
  category = {interaction},
  img      = {RemyChangEtAl2026.png},
}

@String{jrn_NPI_FTR                    = {{Foundations and Trends in Robotics}}}

@String{proc_ACM_HRI                   = {{Proc.\ ACM/IEEE Int.\ Conference on Human Robot Interaction}}}

@String{proc_NIPS-OWA                  = {{Conf.\ on Neural Information Processing Systems - Workshop on Open-World Agents}}}

@InProceedings{AbbeelNg2004,
  Title                    = {Apprenticeship Learning via Inverse Reinforcement Learning},
  Author                   = {Abbeel, P. and Ng, A. Y.},
  Booktitle                = proc_ICML,
  Year                     = {2004}
}

@InProceedings{ZiebartMaasEtAl2008,
  Title                    = {Maximum Entropy Inverse Reinforcement Learning},
  Author                   = {Ziebart, B. D. and Maas, A. and Bagnell, J. A. and Dey, A. K.},
  Booktitle                = proc_AAAI_AAAI,
  Year                     = {2008}
}

@Online{Leurent2018,
  author    = {Leurent, E. and {contributors}},
  title     = {{An Environment for Autonomous Driving Decision-Making}},
  note      = {{Available at }\url{https://github.com/eleurent/highway-env}},
  year      = {2018},
}

@InProceedings{RossGordonEtAl2011,
  author    = {Ross, S. and Gordon, G. J. and Bagnell, A.},
  booktitle = proc_AISTATS,
  title     = {{A Reduction of Imitation Learning and Structured Prediction to No-Regret Online Learning}},
  year      = {2011},
}

@InProceedings{YuYuEtAl2019,
  author    = {Yu, L. and Yu, T. and Finn, C. and Ermon, S.},
  booktitle = proc_NIPS,
  title     = {{Meta-Inverse Reinforcement Learning with Probabilistic Context Variables}},
  year      = {2019},
}

@InProceedings{FinnAbbeelEtAl2017,
  author    = {Finn, C. and Abbeel, P. and Levine, S.},
  booktitle = proc_ICML,
  title     = {{Model-Agnostic Meta-Learning for Fast Adaptation of Deep Networks}},
  year      = {2017},
}

@InProceedings{RamachandranAmir2007,
  Title     = {{Bayesian Inverse Reinforcement Learning}},
  Author    = {Ramachandran, D. and Amir, E.},
  Booktitle = proc_AAAI_IJCAI,
  Year      = {2007},
}

@InProceedings{HoErmon2016,
  author    = {Ho, J. and Ermon, S.},
  booktitle = proc_NIPS,
  title     = {{Generative Adversarial Imitation Learning}},
  year      = {2016},
}

@Article{AbdulhaiJaquesEtAl2023,
  author  = {Abdulhai, M. and Jaques, N. and Levine, S.},
  journal = {{Available at } \url{https://arxiv.org/abs/2208.04919}},
  title   = {{Basis for Intentions: Efficient Inverse Reinforcement Learning using Past Experience}},
  year    = {2023},
}

@Article{SerajLeeEtAl2024,
  author  = {Seraj, E. and Lee, KM. and Zaidi, Z. and Xiao, Q. and Li, Z. and Nascimento, A. and Van Waveren, S. and Tambwekar, P. and Paleja, R. and Das, D. and Gombolay, M.C.},
  journal = jrn_NPI_FTR,
  number  = {2--3},
  pages   = {75--349},
  title   = {{Interactive and Explainable Robot Learning: A Comprehensive Review}},
  volume  = {12},
  year    = {2024},
}

@InProceedings{FilosLyleEtAl2021,
  author    = {Filos, A. and Lyle, C. and Gal, Y. and Levine, S. and Jaques, N. and Farquhar, G.},
  booktitle = proc_ICML,
  title     = {{PsiPhi-Learning: Reinforcement Learning with Demonstrations using Successor Features and Inverse Temporal Difference Learning}},
  year      = {2021},
}

@InProceedings{LiSongEtAl2017,
  author    = {Li, Yunzhu and Song, Jiaming and Ermon, Stefano},
  booktitle = proc_NIPS,
  title     = {{{InfoGAIL}: {Interpretable} imitation learning from visual demonstrations}},
  year      = {2017},
}

@InProceedings{SchrumHedlundEtAl2021,
  author    = {Schrum, M. L. and Hedlund, E. and Gombolay, M. C.},
  booktitle = proc_AAAI_FS,
  title     = {{Improving Robot-Centric Learning from Demonstration via Personalized Embeddings}},
  year      = {2021},
}

@InProceedings{PoddarWanEtAl2024,
  author    = {Poddar, S. and Wan, Y. and Ivison, H. and Gupta, A. and Jaques, N.},
  booktitle = proc_NIPS,
  title     = {{Personalizing Reinforcement Learning from Human Feedback with Variational Preference Learning}},
  year      = {2024},
}

@InProceedings{FuLuoEtAl2018,
  author    = {Fu, J. and Luo, K. and Levine, S.},
  booktitle = proc_ICLR,
  title     = {{Learning Robust Rewards with Adversarial Inverse Reinforcement Learning}},
  year      = {2018},
}

@InProceedings{TorabiWarnellEtAl2018,
  author    = {Torabi, F. and Warnell, G. and Stone, P.},
  booktitle = proc_AAAI_IJCAI,
  title     = {{Behavioral Cloning from Observation}},
  year      = {2018},
}

@Article{SchrumSumnerEtAl2024,
  author  = {Schrum, M.~L. and Sumner, E. and Gombolay, M.~C. and Best, A.},
  journal = jrn_IEEE_TR,
  pages   = {1952--1965},
  title   = {{Maveric: A data-driven approach to personalized autonomous driving}},
  volume  = {40},
  year    = {2024},
}

@InProceedings{SeijenFatemiEtAl2017,
  author    = {van Seijen, H. and Fatemi, M. and Romoff, J. and Laroche, R. and Barnes, T. and Tsang, J.},
  booktitle = proc_NIPS,
  title     = {{Hybrid Reward Architecture for Reinforcement Learning}},
  year      = {2017},
}

@InProceedings{ArjonaMedinaGillhoferEtAl2019,
  author    = {Arjona-Medina, J.~A. and Gillhofer, M. and Widrich, M. and Unterthiner, T. and Brandstetter, J. and Hochreiter, S.},
  booktitle = proc_NIPS,
  title     = {{RUDDER: Return Decomposition for Delayed Rewards}},
  year      = {2019},
}

@Book{SuttonBarto2018,
  author    = {Sutton, R.~S. and Barto, A.~G.},
  year      = {2018},
  title     = {{Reinforcement Learning: An Introduction}},
  publisher = pub_MIT,
}

@InProceedings{JuozapaitisKoulEtAl2019,
  author    = {Juozapaitis, Z. and Koul, A. and Fern, A. and Erwig, M. and Doshi-Velez, F.},
  booktitle = proc_AAAI_IJCAI,
  title     = {{Explainable reinforcement learning via reward decomposition}},
  year      = {2019},
}

@InProceedings{JaquesGuEtAl2017,
  author    = {Jaques, N. and Gu, S. and Bahdanau, D. and Hernández-Lobato, J.~M. and Turner, R.~E. and Eck, D.},
  booktitle = proc_ICML,
  title     = {{Sequence Tutor: Conservative Fine-Tuning of Sequence Generation Models with KL-control}},
  year      = {2017},
}

@InProceedings{OuyangWuEtAl2022,
  author    = {Ouyang, L. and Wu, J. and Jiang, X. and Almeida, D. and Wainwright, C. and L.Mishkin, P. and Zhang, C. and Agarwal, S. and Slama, K. and Ray, A. and Schulman, J. and Hilton, J. and Kelton, F. and Miller, L. and Simens, M. and Askell, A. and Welinder, P. and Christiano, P. and Leike, J. and Lowe R.},
  booktitle = proc_NIPS,
  title     = {{Training language models to follow instructions with human feedback}},
  year      = {2022},
}

@InProceedings{RakellyZhouEtAl2019,
  author    = {Rakelly, K. and Zhou, A. and Quillen, D. and Finn, C. and Levine, S.},
  booktitle = proc_ICML,
  title     = {{Efficient Off-Policy Meta-Reinforcement Learning via Probabilistic Context Variables}},
  year      = {2019},
}

@InProceedings{YeJaques2024,
  author    = {Ye, E. and Jaques, N.},
  booktitle = proc_NIPS-OWA,
  title     = {{An Efficient Open World Benchmark for Multi-Agent Reinforcement Learning}},
  year      = {2024},
}

@article{stable-baselines3,
  author  = {Raffin, A. and Hill, A. and Gleave, A. and Kanervisto, A. and Ernestus, M. and Dormann, N.},
  title   = {Stable-Baselines3: Reliable Reinforcement Learning Implementations},
  journal = {Journal of Machine Learning Research},
  year    = {2021},
  volume  = {22},
  number  = {268},
  pages   = {1-8},
  url     = {http://jmlr.org/papers/v22/20-1364.html}
}

@inproceedings{PalanShevchukEtAl2019,
author = {Palan, M. and Shevchuk, G. and Landolfi, N. and Sadigh, D.},
year = {2019},
month = {06},
pages = {},
booktitle = proc_RSS,
title = {Learning Reward Functions by Integrating Human Demonstrations and Preferences},
}

@inproceedings{EysenbachGuptaEtAl2019,
    author = {Eysenbach, B. and Gupta, A. and Ibarz, J. and Levine, S.},
    title = {Diversity is All You Need: Learning Skills without a Reward Function},
    booktitle = proc_ICLR,
    year = {2019}
}

@inproceedings{ChenPalejaEtAl2020,
    author = {Chen, L. and Paleja, R. and Ghuy, M. and Gombolay, M.},
    title = {Joint Goal and Strategy Inference across Heterogeneous Demonstrators via Reward Network Distillation},
    booktitle = proc_ACM_HRI,
    year = {2020}
}

@inproceedings{TishbyPereiraEtAl1999,
    author = {Tishby, N. and Pereira, F. and Bialek, W.},
    title = {The information bottleneck method},
    booktitle = {{Allerton Conference on Communication, Control and Computing}},
    year = {1999}
}

@inproceedings{AlemiFischerEtAl2019,
    author = {Alemi, A. and Fischer, I. and Dillon, J. and Murphy, K.},
    title = {Deep Variational information bottleneck},
    booktitle = proc_ICLR,
    year = {2019}
}

@inproceedings{ReddyDraganEtAl2020,
    author = {Reddy, S. and Dragan, A. and Levine, S.},
    title = {SQIL: Imitation Learning via Reinforcement Learning with Sparse Rewards},
    booktitle = proc_ICLR,
    year = {2020}
}

@software{gymnasium_robotics2023github,
  author = {Rodrigo de Lazcano and Kallinteris Andreas and Jun Jet Tai and Seungjae Ryan Lee and Jordan Terry},
  title = {Gymnasium Robotics},
  url = {http://github.com/Farama-Foundation/Gymnasium-Robotics},
  version = {1.4.0},
  year = {2024},
}

@inproceedings{HaarjonaZhouEtAl2018-key,
    author ={Haarnoja, T. and Zhou, A. and Abbeel, P. and Levine, S.} ,
    title = {Soft Actor-Critic: Off-Policy Maximum Entropy Deep Reinforcement Learning with a Stochastic Actor},
    booktitle = proc_ICML,
    year = {2018}
}

@phdthesis{Ziebart2010,
    author = {Ziebart, B. D.},
    title = {Modeling purposeful adaptive behavior with the principle of maximum causal entropy},
    school = {Carnegie Mellon University},
    year = {2010}
}

@inproceedings{MnihKavukcuogluEtAl2013,
    author = {Mnih, V. and Kavukcuoglu, K. and Silver, D. and Graves, A. and Antonoglou, I. and Wierstra, D. and Riedmiller, M.},
    title = {Playing Atari with Deep Reinforcement Learning},
    booktitle = proc_NIPS,
    year = {2013}
}

@inproceedings{ChenPaleja2020,
    author = {Chen, L. and Paleja, R. and Gombolay, M.},
    title = {Learning from Suboptimal Demonstration via Self-Supervised Reward Regression},
    booktitle = proc_CoRL,
    year = {2020}
}

@inproceedings{XueChen2025,
    author = {Xue, C. and Chen, L. and Gombolay, M.},
    title = {Better Than Diverse Demonstrators: Reward Decomposition From Suboptimal and Heterogeneous Demonstrations},
    booktitle = jrn_IEEE_RAL,
    year = {2025}
}

@inproceedings{ChenJayanthiEtAl2022,
    author = {Chen, L. and Jayanthi, S. and Paleja, R. and Martin, D. and Zakharov, V. and Gombolay, M.},
    title = {Fast Lifelong Adaptive Inverse Reinforcement Learning from Demonstrations},
    booktitle = proc_CoRL,
    year = {2022}
}
\bibliographystyle{iclr2027_conference}

\appendix
\section{Derivation of Loss Functions}
\label{app:loss}

We aim to maximize the information bottleneck objective \citep{TishbyPereiraEtAl1999}.
\begin{equation}
    I(Z,Y) - \beta I(Z,X)
\end{equation}
Following \cite{AlemiFischerEtAl2019}'s methodology and notation of information bottlenecks, we assume that the joint distribution of $X,Y,Z$ factors into the following:
\begin{equation}
    p(X,Y,Z)=p(Z|X,Y)p(Y|X)p(X)=p(Z|X)p(Y|X)p(X)
\end{equation}
and that $p(Z|X,Y)=p(Z|X)$. 

The mutual information between $Y$ and $Z$ is
\begin{equation}
    I(Z;Y) = \int dy dz \ p(y,z) \log \frac{p(y,z)}{p(y)p(z)} = \int dydz \ \log \frac{p(y|z)}{p(y)} = H(Y) - H(Y|Z)
\end{equation}
where $p(y|z)$ is defined as
\begin{equation}
    p(y|z)=\int dx p(x,y|z) = \int dx \ p(y|x) p(x|z) = \int dx \ \frac{p(y|x)p(z|x)p(x)}{p(z)}.
\end{equation}

However, recovering $p(y|z)$ is intractable so we let $q(y|z)$ be its variational approximation. Since KL divergence is always nonnegative, $D_{KL}(p(Y|Z) || q(Y|Z)) \geq 0$, the following inequality holds:
\begin{equation}
    \int dy \ p(y|z) \log p(y|z) \geq \int dy \ p(y|z) \log q (y|z)
\end{equation}
Thus, $I(Z;Y)$ is bounded from below:
\begin{equation}
    \begin{aligned}
        I(Z;Y) \geq &\int dy dz \ p(y,z) \log \frac{q(y|z)}{p(y)} \\
        &= \int dy dz \ p(y,z) \log q(y|z) - \int dy \ p(y) \log p(y) \\
        &= \int dy dz \ p(y,z) \log q(y|z) + H(Y) \\
        &\approx \mathbb{E}_{y\sim Y,z \sim Z}  \log q(y|z) + H(Y) \\
    \end{aligned}
\end{equation}
Maximizing the mutual information between $Z$ and $Y$ is then equivalent to maximizing $\mathbb{E}_{y\sim Y,z \sim Z} \log q(y|z)$ since $H(Y)$ is constant. We let $q(y|z)$ take the form of a Multivariate Gaussian distribution with fixed variance $\sigma^2$ and assume the network $\theta_q$ that parametrizes this distribution outputs the mean $\hat{\omega}$. Following \citep{SchrumHedlundEtAl2021, SchrumSumnerEtAl2024}, this transforms the mutual information loss into a squared error objective:
\begin{equation}
\begin{aligned}
        q(\omega^{(p)} | s,a,R_s) &= \frac{1}{(\sqrt{2\pi \sigma^2})^d} \exp \bigg (-\frac{1}{2\sigma^2}\sum^d_{i=1}  (\hat{\omega}^{(p)}_i - \omega^{(p)}_i)^2 \bigg ) \\
\end{aligned}
\end{equation}
We $\log$ both sides to derive the following equation.
\begin{equation}
\label{eq:log q}
\begin{aligned}
        \log q(\omega^{(p)}  | s,a,R_s) = \log \frac{1}{(\sqrt{2\pi \sigma^2})^d}  + \log \exp \bigg (-\frac{1}{2\sigma^2}\sum^d_{i=1}  (\hat{\omega}^{(p)}_i - \omega^{(p)}_i)^2 \bigg ) \\
        \log q(\omega^{(p)}  | s,a,R_s) = \log \frac{1}{(\sqrt{2\pi \sigma^2})^d}  + \bigg (-\frac{1}{2\sigma^2}\sum^d_{i=1}  (\hat{\omega}^{(p)}_i - \omega^{(p)}_i)^2 \bigg )
\end{aligned}
\end{equation}
Maximizing the previous equation \eqref{eq:log q} renders out the first term and leaves the squared error term to be minimized. Scaling by $1/d$ leads to \eqref{loss:identity}. 
\begin{equation}
\begin{aligned}
    {\arg \max}_{\theta_q} \log q(\omega | s,a,R_s) = {\arg \max}_{\theta_q} \Biggr \{-\frac{1}{2\sigma^2}\sum^d_{i=1}  \hat{\omega}^{(p)}_i - \omega^{(p)}_i)^2 \Biggr \}\\
    {\arg \max}_{\theta_s} \log q(\omega | s,a,R_s) = {\arg \min}_{\theta_q} \sum^d_{i=1}  (\hat{\omega}_i - \omega_i)^2
\end{aligned}
\end{equation}

Again, following \cite{AlemiFischerEtAl2019}, we now turn our attention toward the compression term. We aim to minimize the mutual information between $X$ and $Z$ defined as
\begin{equation}
\begin{aligned}
    I(X;Z) &= \int dxdz \ p(x,z) \log \frac{p(z|x)}{p(z)} \\
    &= \int dz dx \ p(x,z) - \int dz \ p(z) \log p(z) \\
    &\approx \mathbb{E}_{x\sim p(x)} [D_{KL}(p(z|x) || p(z))] \\
    \end{aligned}
\end{equation}
 We introduce a variational prior $r(z)$ to approximate $p(z)$, the true intractable marginal distribution of the specific rewards. Since the KL divergence between the true marginal and chose prior $r(z)$ is always non negative, we establish a strict upper bound:
\begin{equation}
\label{eq:mi upper}
    \begin{aligned}
            I(X;Z) &\leq \mathbb{E}_{x\sim p(x)} [D_{KL}(p(z|x) ||r(z))] \\
             &= -H(Z|X) - \mathbb{E}_{x,z\sim p(z,x)}[\log r(z)] \big]
    \end{aligned}
\end{equation}

Following our assumption that general rewards constitute the majority of behavior and that specific rewards are then naturally more sparse, we assume $r(z)$ takes the form of a zero-centered Laplace prior: $r(z)=\frac{1}{2b} \exp( -\frac{|z|}{b})$. If instead we assume that the specific behaviors follow a normal distribution, then $r(z)=\frac{1}{\sqrt{2\pi\sigma^2}}\exp(-\frac{z^2}{2\sigma^2})$. The entropy term is constant because our neural network is deterministic. Minimizing \eqref{eq:mi upper} yields the L1 loss.
\begin{equation}
    \begin{aligned}
        \min_{\theta_s} - \mathbb{E}_{x\sim p(x)}\mathbb{E}_{z\sim p(z|x)}[\log r(z)] &\leq \min_{\theta_s} - \mathbb{E}_{x\sim p(x)}\mathbb{E}_{z\sim p(z|x)} \bigg[\log \big (\frac{1}{2b} \big ) - \frac{|z|}{b} \bigg] \\
        &\leq \min_{\theta_s} \mathbb{E}_{x\sim p(x)} \mathbb{E}_{z\sim p(z|x)} [|z|] 
\end{aligned}
\end{equation}

\section{Additional Basis Function Decomposition Results}
\label{app:basis-func}
We consider the following set of equations:
\begin{equation}
    \begin{aligned}
        R_1(x)=\cos(x) + 0.5\sin(x) \\
        R_2(x)=\cos(x) \\
        R_1(x)=\cos(x) - 0.5\sin(x)
    \end{aligned}
\end{equation}
The MSE between $\cos(x)$ and the general reward is $0.0058$.

We consider another set of functions to assess non-additive structures.
\begin{equation}
    \begin{aligned}
        R_1(x)=-0.5\cos(x) \\
        R_2(x)=\cos(x) \\
        R_1(x)=0.5\cos(x) 
    \end{aligned}
\end{equation}
The MSE between $\cos(x)$ and the general reward is $0.1577$. Although the MSE is larger, the recovered reward is still sinusoidal, achieving an angular frequency of $0.99$. The ground truth is 1.

We inject Gaussian noise of increasing variance to assess GRID's robustness on the original set of basis functions: $\{\cos(x)-x, \cos(x), \cos(x) + x\}$.
\begin{table}[h]
    \centering
    \caption{Noise Injection }
    \begin{tabular}{c ccccc}
        $\sigma^2$ & 0 & 0.1 & 0.5 & 1 & 5\\
        \midrule
         $\cos(x)$ MSE & 0.0018 & 0.0027 & 0.0183 & 0.0298 & 0.2228
    \end{tabular}
    \label{tab:noise}
\end{table}
 
\section{Additional Highway Results}
\begin{table}[h]
  \centering
  \caption{Generalist Comparisons in the Highway-Env.}
  \setlength{\tabcolsep}{4pt}
  \resizebox{\textwidth}{!}{%
  \begin{tabular}{l c c c cccc}
    \toprule
    \multirow{2}{*}{\textbf{Method}} & 
    \multirow{2}{*}{\textbf{Reward ($\uparrow$) }} & 
    \multirow{2}{*}{\makecell{\textbf{Episode} \\ \textbf{Length ($\uparrow$)}}} &
    \multirow{2}{*}{\makecell{\textbf{Lane} \\ \textbf{Entropy ($\uparrow$)}}} &
    \multicolumn{4}{c}{\textbf{Lane Occupancy}} \\
    
    \cmidrule(lr){5-8}
    
    & & & & \textbf{Lane 0} & \textbf{Lane 1} & \textbf{Lane 2} & \textbf{Lane 3} \\
    \midrule
    BC & $7.12 \pm 0.24$ & $21.3 \pm 0.6$ & $\mathbf{1.84 \pm 0.01}$ & $0.21 \pm 0.01$ & $0.36 \pm 0.02$ & $0.34 \pm 0.02$ & $0.09 \pm 0.01$ \\
    Pooled AIRL & $12.62 \pm 0.43$ & $34.0 \pm 1.0$ & $1.62 \pm 0.09$ & $0.34 \pm 0.09$ & $0.39 \pm 0.03$ & $0.21 \pm 0.06$ & $0.06 \pm 0.02$ \\
    MSRD & $12.43 \pm 0.28$ & $33.1 \pm 0.7$ & $1.70 \pm 0.11$ & $0.31 \pm 0.08$ & $0.39 \pm 0.04$ & $0.22 \pm 0.03$ & $0.08 \pm 0.03$ \\
    \textbf{GRID-L1 (ours)} & $\mathbf{13.41 \pm 0.18}$ & $\mathbf{35.4 \pm 0.4}$ & $1.77 \pm 0.04$ & $0.13 \pm 0.03$ & $0.49 \pm 0.04$ & $0.26 \pm 0.03$ & $0.12 \pm 0.03$ \\
    \textbf{GRID-L2 (ours)} & $12.68 \pm 0.36$ & $33.6 \pm 0.8$ & $1.74 \pm 0.07$ & $0.16 \pm 0.04$ & $0.48 \pm 0.03$ & $0.25 \pm 0.03$ & $0.11 \pm 0.02$ \\
    \bottomrule
  \end{tabular}%
  }
  \label{tab:gen}
\end{table}

\begin{table}[h]
  \centering
    \caption{Specialist Comparisons in the Highway-Env.}
  \setlength{\tabcolsep}{4pt}
  \resizebox{\textwidth}{!}{%
  \begin{tabular}{l c c c cccc}
    \toprule
    \multirow{2}{*}{\textbf{Method}} & 
    \multirow{2}{*}{\textbf{Reward ($\uparrow$) }} & 
    \multirow{2}{*}{\makecell{\textbf{Episode} \\ \textbf{Length ($\uparrow$)}}} &
    \multirow{2}{*}{\makecell{\textbf{Lane 3} \\ \textbf{Steps ($\uparrow$)}}} &
    \multicolumn{4}{c}{\textbf{Lane Occupancy}} \\
    
    \cmidrule(lr){5-8}
    
    & & & & \textbf{Lane 0} & \textbf{Lane 1} & \textbf{Lane 2} & \textbf{Lane 3} \\
    \midrule
    BC & $7.39 \pm 0.26$ & $36.5 \pm 0.5$ & $\mathbf{34.5 \pm 0.6}$ & $0.01 \pm 0.00$ & $0.02 \pm 0.00$ & $0.03 \pm 0.00$ & $0.94 \pm 0.01$ \\
    Pooled AIRL & $10.23 \pm 0.40$ & $\mathbf{37.3 \pm 0.2}$ & $33.1 \pm 0.6$ & $0.03 \pm 0.01$ & $0.03 \pm 0.01$ & $0.06 \pm 0.01$ & $0.89 \pm 0.02$ \\
    MSRD & $9.91 \pm 0.23$ & $37.2 \pm 0.5$ & $33.7 \pm 0.9$ & $0.02 \pm 0.01$ & $0.03 \pm 0.01$ & $0.05 \pm 0.01$ & $0.91 \pm 0.01$ \\
    \textbf{GRID-L1 (ours)} & $11.37 \pm 0.24$ & $36.2 \pm 0.6$ & $29.8 \pm 1.0$ & $0.01 \pm 0.00$ & $0.05 \pm 0.01$ & $0.12 \pm 0.01$ & $0.82 \pm 0.02$ \\
    \textbf{GRID-L2 (ours)} & $\mathbf{11.43 \pm 0.51}$ & $35.0 \pm 0.4$ & $28.5 \pm 0.3$ & $0.01 \pm 0.00$ & $0.05 \pm 0.01$ & $0.13 \pm 0.01$ & $0.81 \pm 0.01$ \\
    \bottomrule
  \end{tabular}%
  }
  \label{tab:spec}
\end{table}
\begin{figure}[h]
    \centering
    \includegraphics[width=0.7\linewidth]{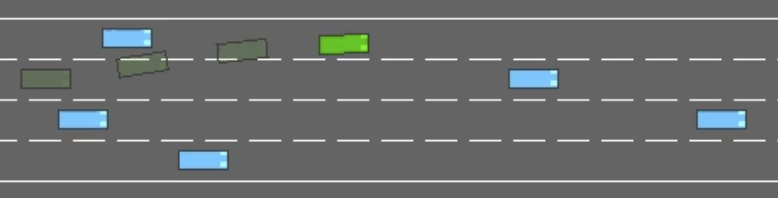}
    \caption{Generalist agent (green vehicle) navigates around other cars (blue vehicles) at high speeds.}
    \label{fig:highway-complex}
\end{figure}

\section{Experiment Setup}
\subsection{Basis Function Decomposition Experiment}

\textbf{Data Generation.} For the synthetic basis function decomposition experiment, we generate 2000 demonstration samples uniformly distributed across the domain $[-4\pi, 4\pi]$.

\subsection{Craftax Experiment}
\textbf{Environment Setup.} To aid in visualizing GRID's disentanglement, we constrain the original Craftax environment. The modifications are as follows:
\begin{itemize}\setlength\itemsep{0.5pt}
    \item The environment map is reduced from a $64 \times 64$ grid to an $8 \times 8$ grid.
    \item The action space for all agents is restricted to \textit{no-op}, \textit{up}, \textit{down}, \textit{left}, \textit{right}, and \textit{do-action}.
    \item The original single-step reward formulation is replaced with a cumulative reward structure.
    \item Adversarial entities (e.g., skeletons, zombies, cows, arrows) are removed.
    \item Agents no longer require tools to retrieve resources such as stone or diamonds.
    \item World generation randomization is largely disabled; only agent spawn positions remain randomized.
    \item Reward weights are modified to ensure each agent prioritizes a distinct resource-gathering task.
    \item Episodes terminate immediately once an agent completes its specific task or is terminated.
\end{itemize}

\textbf{Data Generation.} We generate demonstration data using the IPPO implementation from \citep{YeJaques2024}. To capture a diverse range of skill levels, we collect a total of 2000 demonstration trajectories (500 trajectories each from policy checkpoints at 1, 10, 100, and 200 training iterations). The IPPO actor-critic network utilizes a standard feed-forward architecture comprising three hidden layers of sizes [64, 64, 32] with Tanh activations. The IPPO training hyperparameters are detailed in Table \ref{tab:craftax_ippo}.

\subsection{Highway-Env Experiment}
\textbf{Environment Setup.} We adapt the standard single-agent Highway-Env into a multi-agent environment featuring three agents navigating a four-lane highway. To define the specific behaviors of our heterogeneous population, Agent 0 is rewarded for occupying Lane 0, Agent 1 for Lane 1, and Agent 2 for Lane 2. Lane 3 is left un-preferred by all agents to serve as a holdout specialization target. All lane specialization rewards are valued at $0.1$. Each episode runs for at most 40 steps. 

\textbf{Data Generation.} We modify the PPO implementation from Stable-Baselines3 \citep{stable-baselines3} to support IPPO for our multi-agent setup. Table \ref{tab:highway_ippo} lists the hyperparameters used to train the expert demonstrators. Unlisted parameters are default.

\begin{table*}[t]
\caption{Network architectures and optimization hyperparameters across methods.}
\label{tab:reward_hyperparameters}
\centering
\small
\begin{tabularx}{\linewidth}{l X X}
\toprule
\textbf{Algorithm} & \textbf{Network Architectures} & \textbf{Optimization \& Training Parameters} \\
\midrule
\textbf{GRID} & 
$R_g(s,a)$: MLP [1024, 1024, 1024, 1024], ReLU \newline
$R_s(\omega^{(p)}, s,a)$: MLP [64, 64], ReLU ($\omega^{(p)} \in \mathbb{R}^2$) \newline
$q(R_s^{(p)}, s, a)$: MLP [128, 128] $\to \hat{\omega}^{(p)} \in \mathbb{R}^2$, ReLU & 
Adam lr $1 \times 10^{-3}$, batch size 128, 100 epochs over demo pairs. Targets scaled to $[-1, 1]$. \\
\midrule
\textbf{Per-skill AIRL} & 
Reward $g(s,a)$: MLP [32] \newline
Potential $\phi(s)$: MLP [32, 32] \newline
Generator policy: MLP [64, 64], tanh & 
\textbf{Disc:} Adam lr $1 \times 10^{-3}$, batch 32, 10 steps/round. \newline
\textbf{Gen:} TRPO (KL 0.001, CG 10, batch 100), 200 steps/round. \newline
\textbf{Budget:} 500 rounds (MuJoCo), 200 rounds (Highway). \\
\midrule
\textbf{Pooled AIRL} & 
Reward \& Potential nets same as above. \newline
Generator policy: MLP [64, 64], tanh & 
\textbf{Gen warm start:} BC 2000 steps (lr $3 \times 10^{-4}$). \newline
\textbf{Gen:} TRPO (KL 0.01, batch 100), 1000 steps/round. \newline
\textbf{Disc:} Adam lr $3 \times 10^{-4}$, batch 32, 2 steps/round. 200 rounds. \\
\midrule
\textbf{MSRD} & 
$R_0, \tilde{R}_i$: MLP [32] per net \newline
Potential $\phi_0, \tilde{\phi}_i$: MLP [32, 32] per net \newline
Generator policies: MLP [64, 64], tanh & 
\textbf{Gen warm start:} BC 6000 steps (lr $3 \times 10^{-4}$). \newline
\textbf{Gen:} TRPO (KL 0.02, batch 256, CG 10), 1000 steps/strat/round. \newline
\textbf{Disc:} Adam lr $3 \times 10^{-4}$, batch 64, 1 step/strat/round. \newline
\textbf{Rounds:} Pendulum 150, Hopper 100, Highway 200. \\
\midrule
\textbf{FLAIR} & 
Same architectures as MSRD. & 
Phase 1: Same as MSRD budget. \newline
Phase 2: Same as MSRD + BCD step (lr $3 \times 10^{-5}$, 1 step/strat/round; Highway: 5 steps/strat/round). \\
\midrule
\textbf{REPRESENT} & 
Phase 1: Same architectures as MSRD. \newline
Phase 3 ($R_0, \tilde{R}_i$): MLP [256, 256], ReLU, $\mathbf{W} \sim \mathcal{N}(0, 0.001)$, zero bias. & 
Phase 1: $\epsilon \in \{0, 0, 0, 0, 0, 0.05, 0.1, 0.15, 0.2, 0.25\}$. \newline
Phase 2: $\epsilon \in \{0, 0.05, \dots, 0.95\}$ (20 levels) $\times$ 5 rollouts. \newline
Phase 3: Adam lr $1 \times 10^{-4}$, weight decay $0.1 \sum \|\mathbf{W}\|_2^2$, 3000 iterations, batch 64, 5000 snippets/strat. \\
\bottomrule
\end{tabularx}
\end{table*}


\begin{table*}[t]
\caption{Reward processing and shaping formulas applied before feeding to the generalist policies.}
\label{tab:generalist_rewards}
\centering
\small
\begin{tabularx}{\linewidth}{l X l}
\toprule
\textbf{Method \& Env} & \textbf{Reward Transformation / Blending Formula} & \textbf{Parameters} \\
\midrule
\textbf{GRID (L1/L2)} \newline \textit{MuJoCo} & 
\textbf{1. Scaling:} Min-max scaled to $[-1, 1]$ based on demo $(s, a)$ pairs. \newline
\textbf{2. Blending:} $r' = w \cdot r_{\text{scaled}} + (1 - w) \cdot r_{\text{ood}}$ \newline
\textbf{3. Weight:} $w = \exp(-\max(d - \epsilon, 0) / \sigma)$ \newline
where $d$ is the whitened distance to demo states, and $\epsilon$ is the 99th percentile of the demos' own distances. & 
\begin{tabular}[t]{@{}l@{}}
\textbf{Pendulum:} \\
1-NN distance \\
$r_{\text{ood}} = -10$, $\sigma = 2$ \\
\addlinespace
\textbf{Hopper:} \\
50-NN distance \\
$r_{\text{ood}} = -2$, $\sigma = 4$
\end{tabular} \\
\midrule
\textbf{GRID (L1/L2)} \newline \textit{Highway} & 
Same scaling and blending formula as above. \newline 
$d$ is computed via 50-NN distance on the 25-dim observation. & 
\begin{tabular}[t]{@{}l@{}}
$r_{\text{ood}} = -2$ \\
$\sigma = 4$
\end{tabular} \\
\midrule
\textbf{Baselines} \newline \textit{MuJoCo} & 
Raw learned reward passed directly to SAC (no scaling). & 
N/A \\
\midrule
\textbf{Baselines} \newline \textit{Highway} & 
Min-max mapped to $[0, 1]$ on demo pairs, then clipped. \newline
\textit{(Ensures staying alive never pays less than crashing)} & 
N/A \\
\bottomrule
\end{tabularx}
\end{table*}

\begin{table*}[t]
\caption{Generalist policy architectures and optimization hyperparameters.}
\label{tab:generalist_hyperparameters}
\centering
\small
\begin{tabularx}{\linewidth}{l X X}
\toprule
\textbf{Algorithm \& Task} & \textbf{Network Architecture} & \textbf{Optimization \& Training Parameters} \\
\midrule
\textbf{6.1 SAC} \newline (MuJoCo) & 
\textbf{Algorithm:} SAC (\texttt{SBX}) \newline
\textbf{Actor:} MLP [256, 256], ReLU \newline
\textbf{Critics (2):} MLP [256, 256], ReLU & 
\textbf{Opt:} lr $3 \times 10^{-4}$, batch 256, buffer $10^6$. $\tau = 0.005, \gamma = 0.99$. \newline
\textbf{Routine:} Train freq 1, 1 grad step/env step. Learning starts at 100. Auto entropy tuning. \newline
\textbf{Env:} 1000-step eps, no fall termination. \newline
\textbf{Budget:} Pendulum 100k, Hopper 1M steps. \\
\midrule
\textbf{6.2 BC Generalist} \newline (MuJoCo) & 
\textbf{Actor:} Squashed-Gaussian MLP [256, 256] (\texttt{SBX}) & 
\textbf{Objective:} Max likelihood on all pooled demo pairs. \newline
\textbf{Opt:} Adam lr $3 \times 10^{-4}$, batch 256. \newline
\textbf{Budget:} 20k steps. \\
\midrule
\textbf{6.3 DQN} \newline (Highway) & 
\textbf{Algorithm:} DQN (\texttt{SB3}) \newline
\textbf{Q-Network:} MLP [256, 256] & 
\textbf{Opt:} lr $5 \times 10^{-4}$, batch 32, buffer 15k, $\gamma = 0.95$. Target update every 50 steps. \newline
\textbf{Routine:} Train freq 1, 1 grad step/env step. Learning starts at 200. \newline
\textbf{Exploration:} $\epsilon$-greedy decays $1.0 \to 0.05$ (first 10\%). \newline
\textbf{Env:} 32 subprocess envs. Random learner car. Ends on crash or 40 steps. Budget: 100k steps. \newline
 \\
\midrule
\textbf{6.4 BC Generalist} \newline (Highway) & 
\textbf{Q-Network:} MLP [256, 256] \newline
(Treated as a 5-way action classifier) & 
\textbf{Objective:} Cross-entropy on all 1200 pooled demo pairs. \newline
\textbf{Opt:} Adam lr $3 \times 10^{-4}$, batch 256. \newline
\textbf{Budget:} 20k steps. \\
\midrule

\bottomrule
\end{tabularx}
\end{table*}

\begin{table}[htbp]
\centering
\caption{IPPO Training Hyperparameters: Craftax Data Generation}
\begin{tabular}{ll}
\toprule
\textbf{Hyperparameter} & \textbf{Value} \\
\midrule
Training iterations per agent   & 200 \\
Parallel environments           & 200 \\
Minibatches                     & 16 \\
Learning rate                   & $2.5 \times 10^{-4}$ \\
Entropy coefficient             & 0.02 \\
Max gradient norm               & 1.0 \\
Discount factor ($\gamma$)      & 0.99 \\
GAE parameter ($\lambda$)       & 0.95 \\
Clip coefficient                & 0.2 \\
Value function coefficient      & 0.5 \\
Auxiliary loss coefficient      & 0.1 \\
\bottomrule
\end{tabular}
\label{tab:craftax_ippo}
\end{table}

\begin{table}[htbp]
\centering
\caption{IPPO Training Hyperparameters: Highway-Env Data Generation}

\begin{tabular}{ll}
\toprule
\textbf{Hyperparameter} & \textbf{Value} \\
\midrule
Learning rate & $3 \times 10^{-4}$ \\
Steps per rollout ($n\_steps$) & 2048 \\
Batch size & 256 \\
Epochs & 10 \\
Discount factor ($\gamma$) & 0.95 \\
Entropy coefficient & 0.05 \\
\bottomrule
\end{tabular}
\label{tab:highway_ippo}
\end{table}

\begin{figure}[h]
    \centering
    \includegraphics[width=0.5\linewidth]{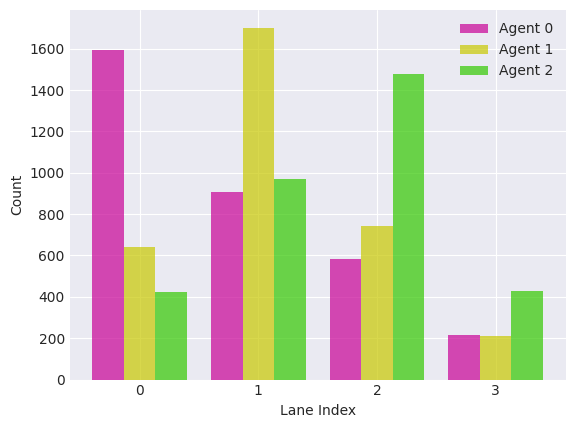}
    \caption{IPPO Agents Lane Counts over 100 Episodes. Average Episode Length=$33.0$}
    \label{fig:ippo-counts}
\end{figure}



\subsection{MuJoCo Experiments}
We use Gymnasium InvertedPendulum-v5 and Hopper-v5 for our continuous control tasks. Both tasks are run for 1000 steps and use the default environment reward as the shared reward. The total reward is
\begin{equation}
    r(s,a)=r_{\text{environment}}(s,a) + 0.5\cdot r_{\text{DIAYN}}(s)
\end{equation}
Pendulum is trained for 100k steps and hopper is trained for 1M steps. 10 demonstrations are collected for each.

\begin{table}[h]
\caption{Hyperparameter settings for Soft Actor-Critic (SAC) and the DIAYN Skill Discriminator.}
\label{tab:hyperparameters}
\centering
\begin{tabular}{ll}
\toprule
\textbf{Hyperparameter} & \textbf{Value} \\
\midrule
\multicolumn{2}{c}{\textbf{Soft Actor-Critic (SAC)}} \\
\midrule
Architecture & MLP [256, 256] \\
Activation & ReLU \\
Learning Rate & $3 \times 10^{-4}$ \\
Replay Buffer Size & $10^6$ \\
Batch Size & 256 \\
Target Smoothing Coefficient ($\tau$) & $0.005$ \\
Discount Factor ($\gamma$) & $0.99$ \\
Train Frequency & 1 \\
Gradient Steps & 1 per env step \\
\midrule
\multicolumn{2}{c}{\textbf{Skill Discriminator}} \\
\midrule
Architecture $q(z|s)$ & MLP (width 64, depth 2) \\
Optimizer & Adam \\
Learning Rate & $3 \times 10^{-4}$ \\
Batch Size & 256 \\
Replay Buffer Size & $10^5$ \\
Gradient Steps & 1 per env step \\
\bottomrule
\end{tabular}
\end{table}

\end{document}